# Improving the interaction of Older Adults with Socially Assistive Robots for Table setting

Samuel Olatunji, Noa Markfeld, Dana Gutman, Shay Givati,
Vardit Sarne-Fleischmann, Tal Oron-Gilad, Yael Edan

Ben-Gurion University of the Negev, Beer-Sheva, Israel

**Abstract.** This study provides user-studies aimed at exploring factors influencing the interaction between older adults and a robotic table setting assistant. The influence of level of automation (LOA) and level of transparency (LOT) on the quality of the interaction was considered. Results revealed that the interaction effect of LOA and LOT significantly influenced the interaction. A lower LOA which required the user to control some of the actions of the robot influenced the older adults to participate more in the interaction when the LOT was low compared to situations with higher LOT (more information) and higher LOA (more robot autonomy). Even though, the higher LOA influenced more fluency in the interaction, the lower LOA encouraged a more collaborative form of interaction which is a priority in the design of robotic aids for older adult users. The results provide some insights into shared control designs which accommodates the preferences of the older adult users as they interact with robotic aids such as the table setting robot used in this study

## 1 Introduction

Robots with improved capabilities are advancing into prominent roles in assisting older adults in performing daily living tasks such as cleaning, dressing, feeding (Shishehgar, Kerr and Blake, 2018)(Shanee *et al.*, 2018). This has to be done with careful consideration for the strong desire of these older adults to maintain a certain level of autonomy while carrying out their tasks of daily living, even if the robot provides the help they require (Wu *et al.*, 2016). Furthermore, the robot's involvement should not drive the older adult to boredom, sedentariness or loss of skills relevant to daily living due to prolonged inactivity (Beer, Fisk and Rogers, 2014). A possible solution is shared control where the user preferences are adequately considered as the robot's role and actions are being defined during the interaction design. This ensures that the older adults are not deprived of their cherished independence (Zwijsen, Niemeijer and Hertogh, 2011).

This study, proposes a shared control strategy using levels of automation (LOA) which refers to the degree to which the robot would perform particular functions in its defined role of assisting the user in a specific task (Parasuraman, Sheridan and Wickens, 2008). The aim is to ensure high quality collaboration between the older adult and the robot in accomplishing desired tasks, without undermining the autonomy, preferences and satisfaction of the older adult.

To ensure transparency of the robot's role at all times the LOA implementation is reflected in the ways through which the users interact with the robots. Transparency in this context is the degree of task-related information provided by the robot to the older adults to keep them aware of its state, actions and intentions of the robot (Chen *et al.*, 2018). The content of this information provided by the robot can be graded according to the detail, quantity and type of information as mirrored in Endsley's situation awareness (SA) study (Endsley, 1995) and Chen *et al.*'s SA-based Transparency model (Chen *et al.*, 2014). It is essential that the level of transparency (LOT) of the information being presented to the older adults conforms with their perceptual and cognitive peculiarities such as the processing and interpretation of the information provided by the robot (Rogers and Mitzner, no date; Mitzner *et al.*, 2015; Feingold Polak *et al.*, 2018). Existing studies reveal that the information presented to the users significantly influences their comprehension of the robot's behavior, performance and limitations (Dzindolet *et al.*, 2003; Lyons, 2013; Chen *et al.*, 2014). This information facilitates the users' knowledge of the automation connected to the task (Endsley, 2017). This affects the users' understanding of their role and that of the robot in any given interaction (Lyons, 2013; Chen *et al.*, 2014; Doran, Schulz and Besold, 2017; Hellström and Bensch, 2018).

Some studies explored the presentation of information in different technological aids for older adults were given by (Cen/Cenelec, 2002; Fisk *et al.*, 2009; Mitzner *et al.*, 2015). These studies, provided recommendations serving as design guidelines for information presented in various modes such as visual, audial or haptic information. These recommendations were for information presented on digital mobile applications, webpages, equipment, and other facilities through which older adults would interact with their environment. They were not specific to information presented by robots to the older adults. It was recommended in those studies that sufficient user studies should be conducted in specific domains to identify suitable design parameters that would be fitting to specific applications (Cen/Cenelec, 2002; Fisk *et al.*, 2009; Mitzner *et al.*, 2015; van Wynsberghe, 2016).

This current study aims to explore how LOA and LOT influences the quality of interaction (QoI) between the older adults and the assistive robot in a shared task of table setting. The QoI is a construct in this paper which entails the fluency, understanding, engagement and comfortability during the interaction

## 2 Methods

### 2.1 Overview

A table setting task performed by a robotic arm was used as the case study. The robot had to pick up a plate, a cup, a fork and a knife and to place them at specific positions on the table. The user operated the robot in two levels of automation. In the high LOA condition, the robot operated autonomously. The user could only start and stop the robot's operation by pressing a specific button. In the low LOA condition, the user could still start and stop the robot at will, but the robot required the user's consent before setting each item. The robot asked the user through a GUI which item to bring and the user was required to respond before the robot could continue its operation.

Two conditions utilizing different levels of transparency (LOT) were compared for two different levels of the robot's automation: high and low (Table 1). Information was given by the robot in visual form through a GUI on an adjacent screen where the LOT manipulation was effected (Figure 1). The two conditions differed by the amount of details provided by the robot. The low level of information included text messages that specified the status of the robot by indicating **what** it was doing (e.g. bringing a plate, putting a fork, etc.), while the high level of information included also the **reason** for this status (i.e. I'm bringing the plate since you asked me, etc.)

**Table 1.** Experimental Conditions.

|  |  | LOA | |
|---|---|---|---|
|  |  | *Low* | *High* |
| **LOT** | *Low* | Condition 1 – LL<br>User instructs the robot using the GUI and receives information about **what** the robot is doing in each stage. | Condition 3 – LH<br>Robot operates automatically. In each stage user receives information about **what** the robot is doing. |
|  | *High* | Condition 2-HL<br>User instructs the robot using the GUI and receives information about what the robot is doing and the reason for it in each stage. | Condition 4-HH<br>Robot operates automatically. In each stage user receives information about **what** the robot is doing and the **reason** for it. |

### 2.2 Apparatus

A KUKA LBR iiwa 14 R820 7 degrees of freedom robotic arm equipped with a pneumatic gripper was used (Figure 1). The tasks were programmed using python and executed on the ROS (Schaefer, 2015) platform.

In order to instruct the robot and to present the information received by the robot a graphical user interface (GUI) was used on PC screen, which was located on a desk to the left of the user (see Figure 1).

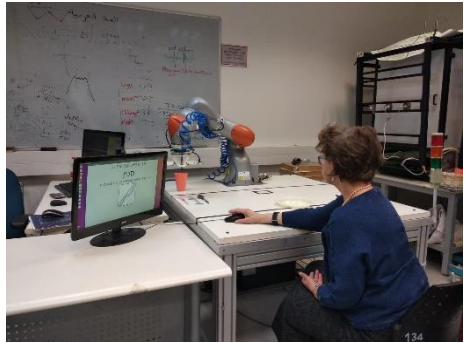

**Fig. 1.** A participant using the GUI to instruct the robot.

### 2.3 Participants

Fourteen older adults (8 Females, 6 Males) aged 62-82 (mean 69.8) participated in the study. Participants were recruited through an ad which was publicized electronically. They were healthy individuals with no physical disability who came independently to the lab. Each participant completed the study separately at different timeslots, so there was no contact between participants.

### 2.4 Experimental Design

The experiment was set with a mixed between and within subject design with the LOA modes as the between subject variable, and the LOT as the within subject variable.

Participants were assigned randomly to one of the two LOA conditions. All participants completed the same table setting task for both levels of transparency. The order of the two tasks was counterbalanced between participants, to accommodate for potential bias of learning effects, boredom or fatigue.

### 2.5 Performance measures

Initially, participants completed a pre-test questionnaire which included the following: demographic information, and a subset of questions from the Technology Adoption Propensity (TAP) index (Ratchford and Barnhart, 2012) to assess their level of experience with technology and from the Negative Attitude toward Robots Scale (NARS) (Syrdal *et al.*, 2009) to assess their level of anxiety towards robots.

Objective measures that were collected during each session are interaction-related variables such as fluency, engagement, understanding and comfortability. Subjective measures were assessed via questionnaires. Participants completed a short post-session questionnaire after each session and a final questionnaire at the end of the two sessions (Figure 1) to evaluate subjective measures. The post-session questionnaire used 5-point Likert scales with 5 representing "Strongly agree" and 1 representing "Strongly disagree". The final questionnaire related to the difference between both sessions.

## 2.6 Analysis

A two-tailed General Linear Mixed Model (GLMM) analysis was performed to evaluate for a positive or negative effect of the independent variables. The user ID was included as a random effect to account for individual differences. LOA and LOT were utilized as fixed factors while all objective and subjective variables representing 'Quality of Interaction' (QoC) were used as dependent variables.

## 3 Results

### 3.1 Demographics and Attitude towards Technology

There was an equal distribution of participants within the two groups. On a scale of 1 (strongly disagree) to 5 (strongly agree), the TAP index reveals that most of the participants are optimistic about technology providing more control and flexibility in life (*mean = 3.86, SD=1.17*). It was also observed that over 75% of the participants like to learn the use of new technology (*mean=3.93, SD=1.07*) and feel comfortable communicating with robots (*mean= 3.43, SD=1.50*). The majority (80%) did not have negative feelings about situations in which they have to interact with a robot (*mean = 4.14, SD = 0.86*).

### 3.2 Quality of Interaction

There was a significant difference between the LOA-LOT manipulation as conditions (*p=0.033*). The effect of the manipulation was significant on the robot's idle time (*p=0.009*), functional delay (*p<0.001*), human idle time (*p=0.005*), the gaze on the robot (*p=0.021*), perception of safety (*p=0.042*) and overall interaction time (*p=0.007*). The effect of the manipulation was not significant on the gaze on the GUI where the robot provided feedback (*p=0.134*) and the perception of safety (*p=0.087*).

### 3.3 Fluency

Fluency was represented by the idle time of the robot, functional delay and overall time spent on the task. The LOA was significant on the robot's idle time (*mean = 122.54, SD = 59.70, $F(1, 24) = 9.97$, p=0.004*) with the high LOA (*mean=88.85, SD=2.48*) having a lower robot idle time compared to low LOA (*mean 156.21, SD=70.38*). The LOT was not significant as a main effect but there was a significant effect in the interaction between the LOA and LOT (*$F(4, 24) = 44.2$, p<0.001*) as depicted in Fig. 3. In terms of delay (*mean = 12.86, SD = 13.87*), the LOA was significant (*$F(1, 24) = 14.48$, p=0.001*). The low LOA had more delays (*mean=20.85, SD=15.99*) than high LOA (*mean=4.87, SD=13.87*). The LOT was not significant (*$F(1, 24) = 2.04$, p=0.17*). There was also no interaction effect of the LOA and LOT on the delays (*$F(1, 24) = 1.49$, p=0.23*). The duration of the experiment with low LOA (*mean=239.21, SD=74.41*) were longer than that with high LOA (*mean=158.53, SD=66.17*). This was also statistically significant (*mean = 198.53, SD = 66.17, $F(1, 24) = 15.42$, p=0.001*). The results therefore suggest that high LOA influenced more fluency in the interaction than low LOA.

### 3.4 Engagement

The duration of the gaze on the robot was significantly affected by LOA (*mean = 155.64, SD = 34.51, p=0.006*). Participants in low LOA (*mean=175.57, SD=34.77*) gazed on the robot more than participants in high LOA (*mean=135.71, SD=20.22*). The interaction between LOA and LOT on the time participants gazed on the robot was significant (*$F(1,24)=7.83, p=0.01$*). Participants in low LOA (*mean=35.50, SD=17.81*) were also more significantly focused on the GUI (*mean = 27.01, SD = 19.60, p =0.037*) than participants in high LOA (*mean=18.643, SD=18.10*). The interaction between LOA and LOT was significant regarding the focus on GUI (*$F(1, 24) = 4.48, p=0.045$*). The effect of LOA on the human's active time was also significant (*mean = 16.39, SD = 16.62, p<0.001*) with low LOA (*mean = 31.07, SD=10.47*) keeping the human more active than the high LOA (*mean=1.71, SD=0.82*). There was an interaction effect between the LOA and LOT (*$F(1, 24) = 47.28, p<0.001$*)

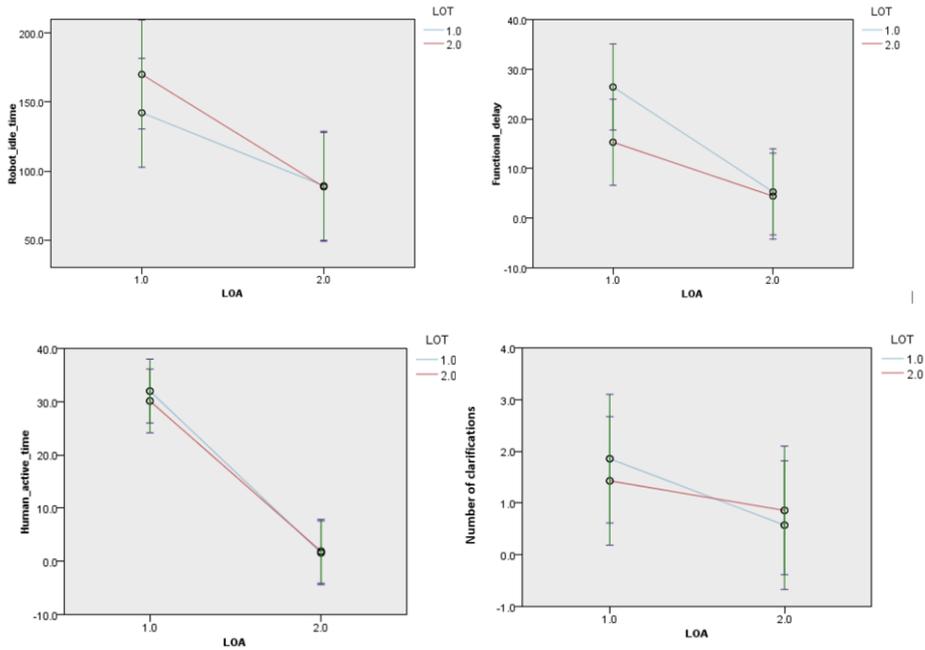

**Fig. 2.** Interaction effect of LOA and LOT on various some QoI variables

### 3.5 Understanding

The was no significant difference in the number of clarifications made by the participants during the interaction (*mean=1.18, SD=1.59, p=0.124*) as a result of the LOA manipulation. The participants seemed to understand the status of the interaction and actions of the robot in both LOA and LOT modes (*$F(1, 24) = 2.27, p=0.15$*). Only a few participants asked for clarification at the low LOA (*mean=1.64, SD=1.95*) and high LOA modes (*mean=0.71, SD=0.99*). However, in terms of reaction time of the participants as the robot interacted with them, the LOA was significant (*mean = 12.86, SD = 13.87, p=0.001*). The participants spent more time observing and processing the information the robot was presenting to them as consent in the low LOA (*mean=20.85, SD=15.99*) compared to the high LOA (*mean=4.87, SD=13.87*).

### 3.6 Comfortability

The effect of the LOA and LOT did not influence the heart rate of the participants. But it was also not significant on the comfortability of the participants with regards to their perception of safety of the robot (*mean = 2.54, SD = 0.58, p =0.48*). However, it was observed that Participants in low LOA moved much closer to the robot which represented more comfortability with it than participants in high LOA which sat further away from the robot.

## 4 Discussion and Conclusion

Most of the participants were comfortable interacting with a robot. The results revealed that the quality of interaction, as measured via fluency, engagement, understanding and comfortability of the interaction was influenced mainly by the interaction of LOA and LOT. The LOT alone had less influence. It was observed during the experiments that participants seem to prefer less information (low LOT) when they were more active with the robot (high LOA). This agrees with the findings in (Chen *et al.*, 2018) where differences were not found in the transparency level that included only status information and reason without LOA involved. In this study where the level of involvement of the participant varies with the LOA, it is noteworthy that the LOT preferred is influenced by the LOA the robot is operating in. The participants seem to prefer the robot to give more information when it operated more autonomously.

This corroborates the characteristics of the visuospatial sketchpad (VSSP) working principle as modelled by Baddeley (Baddeley *et al.*, 1975; Baddeley, 1986, 1990). It suggests a dissociation within the VSSP, between active operations such the movement of the robot and a passive store of information, such as that displayed on the GUI (Bruyer and Scailquin, 1998). There is more cognitive demand on the participants when actively involved with the robot which suggests a reason why the participants seem to prefer less information to process. This is in contrast to the scenario where the robot was more autonomous, with less cognitive demand on the participant.

Future research should advance a longitudinal study, to increase familiarity with the robot operation and overcome the suspected naivety effect (Shah and Wiken, 2011; Kirchner and Alempijevic, 2012) of the older adults with the robot. We expect that the more the older adults get familiar with the operation of the robot, their level of trust in the robot may change and thus cause a change in their LOT demands as well.

According to the participants' recommendations more awareness might be improved through voice feedback. This possibility is also supported by the suggestion of (Sobczak-Edmans *et al.*, 2016) indicating that some form of verbal representation of information supports visual representations. This should be explored in future work to improve the shared control of the older adult with the table setting robot.

Previous research in human robot collaboration discovered the effectiveness of coordination in team performance as presented in (Shah and Wiken, 2011). Our work further presents the potential of LOA in improving quality of interaction. This is reflected in the various objective measures taken for engagement, fluency, degree of involvement and comfortability with the robot where the LOA effect was significant. The low LOA enabled the participant to interact more with the robot by selecting the specific item that the robot should pick up and the order of arrangement. This inspired greater collaboration with the robot. It enhanced the concept of shared control where the user is more involved in the decisions and control of the robot's operations. This is very critical to ensure that the human keeps the use of various motor functions of the arm so as not to lose skills or functionality of the muscles (Wu *et al.*, 2014). This corresponds with the "use it or lose it" logic presented by (Katzman, 1995) in their study of older adult lifestyle.

Most studies which included some form of adaptive coordination to improve the collaboration between the robot and the user (Huang, Cakmak and Mutlu, 2015; Someshwar and Edan, 2017) tried to reduce the completion time of the task. There was a trade off in this current study regarding degree of involvement and time to complete task i.e., at a higher degree of user involvement, more time was spent to complete the task. It is noteworthy that the peculiar nature of the target population demands that it is important to ensure user involvement to avoid idleness and other negative outcomes of sedentariness. Moreover, most participants expressed enjoyment, and pleasure as they interacted with the robot, which suggests other reasons for the longer interactive time. This can therefore be considered as a positive outcome of the interaction and a favorable contribution to improve shared control in human-robot interaction scenarios such as this.

## Acknowledgments

This research was supported by the EU funded Innovative Training Network (ITN) in the Marie Skłodowska-Curie People Programme (Horizon2020): SOCRATES (Social Cognitive Robotics in a European Society training research network), grant agreement number 721619 and by the Ministry of Science Fund, grant agreement number 47897. Partial support was provided by Ben-Gurion University of the Negev through the Helmsley Charitable Trust, the Agricultural, Biological and Cognitive Robotics Initiative, the Marcus Endowment Fund, the Center for Digital Innovation research fund, the Rabbi W. Gunther Plaut Chair in Manufacturing Engineering and the George Shrut Chair in Human Performance Management.